# Kurdish Handwritten Character Recognition using Deep Learning Techniques

**Rebin M. Ahmed[1], Tarik A. Rashid[2]\*, Polla Fattah[3], Abeer Alsadoon[4,5], Nebojsa Bacanin[6], Seyedali Mirjalili [7,8] [9]S.Vimal, [10]Amit Chhabra**

**Abstract**: Handwriting recognition is one of the active and challenging areas of research in the field of image processing and pattern recognition. It has many applications that include: a reading aid for visual impairment, automated reading and processing for bank checks, making any handwritten document searchable, and converting them into structural text form, etc. Moreover, high accuracy rates have been recorded by handwriting recognition systems for English, Chinese Arabic, Persian, and many other languages. Yet there is no such system available for offline Kurdish handwriting recognition. In this paper, an attempt is made to design and develop a model that can recognize handwritten characters for Kurdish alphabets using deep learning techniques. Kurdish (Sorani) contains 34 characters and mainly employs an Arabic\Persian based script with modified alphabets. In this work, a Deep Convolutional Neural Network model is employed that has shown exemplary performance in handwriting recognition systems. Then, a comprehensive dataset was created for handwritten Kurdish characters, which contains more than 40 thousand images. The created dataset has been used for training the Deep Convolutional Neural Network model for classification and recognition tasks. In the proposed system, the experimental results show an acceptable recognition level. The testing results reported a 96% accuracy rate, and training accuracy reported a 97% accuracy rate. From the experimental results, it is clear that the proposed deep learning model is performing well and is comparable to the similar model of other languages' handwriting recognition systems.

Rebin M. Ahmed
rebin.mohammed@tiu.edu.iq

Tarik A. Rashid
tarik.ahmed@ukh.edu.krd

Polla Fattah
pollaeng@gmail.com

Mohammed Kamal Majeed
mohammed.kamal@tiu.edu.iq

Abeer Alsadoon
alsadoon.abeer@gmail.com

Nebojsa Bacanin
nbacanin@singidunum.ac.rs

Seyedali Mirjalili
ali.mirjalili@gmail.com

S.Vimal
svimalphd@gmail.com

Amit Chhabra
amit.cse@gndu.ac.in

[1]\* IT Department, Faculty of Applied Science, Tishk International University, Erbil, Iraq

[2]\* Computer Science and Engineering Department, University of Kurdistan Hewler, Erbil Iraq

[3] Software and Informatics Engineering, Salahaddin University-Erbil-, Erbil, Iraq

[4] School of Computing and Mathematics, Charles Sturt University, Sydney, Australia

[5] Information Technology Department, Asia Pacific International College (APIC), Sydney, Australia

[6] Singidunum University, Belgrade, Serbia

[7] Centre for Artificial Intelligence Research and Optimisation, Torrens University, Australia

[8] Yonsei Frontier Lab, Yonsei University, Seoul, Korea

[9] Department of Artificial Intelligence and Data Science, Ramco Institute of Technology, North Venganallur Village, Rajapalayam - 626 117 Virudhunagar District Tamilnadu, India

[10] Department of Computer Engineering and Technology, Guru Nanak Dev University, Amritsar-India

# 1. Introduction

Currently, handwritten character recognition is increasing in demand and popularity due to its potential application areas, which would reduce the data entry tasks, save an unimaginable time when filling, searching for the form in handwritten documents, and much more. The work of this study shows the consideration of the following problems:

1) Character recognition is a challenging and broad area for research, and still, improvement is required because the accuracy of the existing models is nowhere near human capabilities. Furthermore, handwriting character recognition is even more complex and challenging to be processed due to the nature of handwriting itself (Zhang Xu-Yao, Yoshua Bengio, Cheng-Lin Liu, 2017).
2) When researching Kurdish handwriting, some issues need to be focused on, and an efficient recognition system with a suitable mechanism should be worked on to manage it. These issues make handwritten Kurdish characters more challenging and complicated than Latin character recognition because:
   i. The characters' format, pattern, and shape are similar to Arabic, Persian, Urdu, and other similar languages.
   ii. They include a massive character collection with more curves, loops, and other different details of the characters.
   iii. Many character sets are pretty similar in shape.
3) Designing and implementing a multiclass classification system requires a comprehensive dataset for training and evaluating the model. Unfortunately, at the moment, there is no handwritten character dataset available for the Kurdish language.

The primary aspiration and motive behind this research are to provide bases for the Kurdish language's handwritten character recognition. We saw a great need since we have thousands of handwritten materials, from government documents to court cases, historical and archeological documents, poetry and classical literature, and many more.

The motivation of this dissertation considers as follows:

1) Provide bases for the Kurdish Language handwritten character recognition. (written in Persian-Arabic scripts)
2) Create a dataset of handwritten Kurdish characters.

The theoretical work of this study has involved some good contributions in developing a much higher understanding of character recognition and deep learning concepts, models, and procedures. This study aimed at developing a convolutional neural network model for character recognition as well as a new dataset of Kurdish handwritten letters, like the following:

1) Producing a large dataset of nearly 41 thousand Kurdish handwritten letters that's the first of its kind for the Sorani dialect and making it available for the research community.
2) Creating a convolutional neural network architecture/model that can classify and recognize Kurdish handwritten characters.

Two different writing systems have been used for the Kurdish language. It is written in Arabic script in Iraq and Iran, and in Latin script in Turkey, Syria, and Armenia. Our research focuses only on Sorani (Central Kurdish), and it is widely assumed that about 9 to 10 million people speak Sorani in both Iran and Iraq. It is primarily written using a modified Arabic script with 34 characters. Since Sorani uses a script that is written using a cursive style, which means certain characters might not take the same forms based on the location of that character within a particular word, our dataset includes only isolated characters.

This paper can be directed as follows: Section 2 presents previous works. Section 3 is divided into two main subsections. 3.1 manifests the description of the dataset, and 3.2 shows the designing process of the model architecture. Section 4 is also divided into two parts. 4.1 explains the training process, and 4.2 shows the different test and evaluation methods applied. Finally, section 5 summarizes the process, discussions, and future works.

## 2. Literature Review

There were very few studies on character recognition for the Kurdish language in the past, and the published research is either for printed characters or online recognition. Kurdish writing is among the most complex and sophisticated structures (Bayan Omar Mohammed, 2013), starting from the character's old eastern line drawing format. The character's form is rather generic and resembles other characters. Nonetheless, each character remains distinctive. Such characteristics may be generalized as handwriting.

Research by Bayan Omar Mohammed (2013) appears to be one of the earliest investigations into Kurdish handwriting character recognition. Her study intends to focus on the Kurdish language characters to extract geometric moment characteristics for the handwriting recognition shape characters, and the geometric moment's features are being completed thoroughly. Also, by checking and investigating it efficiently, the presence of single features will be legitimized; hence, the idea of applying the Invariant Discretization. Injecting the solo performance into the system by injecting various problems into a unique feature or standard performance for a single feature is accomplished through Invariant Discretization support. Although the core focus is on the isolated handwritten Kurdish characters, the benefit of Invariant Discretization increases when the recognition level for each handwriting can be increased. The use of Fuzzy logic classifiers is being investigated to get to know isolated Kurdish handwritten characters.

Behnam Zebardast, Isa Maleki, and Awat Maroufi (2014) used neural networks as one of the various character recognition methods. Artificial neural networks could have been successful according to the parallel process flexibility and learning capacity for particular applications such as pattern recognition, and it's an acceptable approach to understanding the Kurdish alphabet. The Kurdish language has two manuscripts which are the Latin and Arabic alphabets. Their study investigated artificial neural networks, particularly Multilayer Perceptron (MLP), using a backpropagation learning algorithm for classifying the Kurdish language (Latin scripts). Performance in the training stage is 85.1535 percent and in the evaluation, the stage is 81.2677 percent. The accuracy recognition rate is 85.1535 percent in the training phase and 81,2677 percent in the test phase. The model suggested in this work is useable with Latin letters for all languages.

A hybrid implementation of the Harmony Search algorithm and Hidden Markov Model for character recognition was introduced by Rina D.Zarro and Mardin A.Anwer (2017). Similar to most earlier works, the Markov model is used as an intermediary classifier rather than the main classification method. The HMM model is used to classify characters into smaller categories by structural elements based on a shared directional matrix. This approach minimized computation duration in the later recognition stage. As a fitness function, the harmony search algorithm was a recognizer, which used representative and systematic patterns of movement; besides, the actual function is used to lower the matching rate through fitness function measurements. This technique was evaluated on a dataset consisting of 4500 words along with 21,234 typescripts in various places. The system achieved a 93.52% successful recognition rate.

Rasty Yaseen and Hossein Hassani (2018) Proposed an OCR platform to segment and classify Kurdish texts in Persian/Arabic script. Their solution is to adjust and develop the proposed methods for Farsi (Persian) and Arabic, particularly segmentation methods based on contour labeling. They did several tests on fonts, text size, and image resolutions. As a result, they achieved an average of 90.82% accuracy in the recognition rate. However, the accuracy was lower for a few cases, depending on the font style, size, or having mixed fonts on a document.

More recently, Mohammed, Twana, Ahmed, and Al-Sanjary, have published the Kurdish Offline Handwritten Text Dataset (Mohammed, T.L., Ahmed, A.A. and Al-Sanjary, O.I., 2019), which has a total of 4304 manuscripts written by 1076 volunteers in different ages, genders, and levels of education. The 4304 manuscripts had a total of 17466 lines of text. Unfortunately, at the time of writing, there is no published research on the accuracy or the performance of the dataset mentioned.

Furthermore, new on printer character recognition has been published (Saman Idrees, Hossein Hassani, 2021), and researchers have studied Tesseract LSTM, which is a popular Optical Character Recognition (OCR) engine that has been trained to recognize many popular languages. However, it needs very good resources and training data, if not available the result will become less accurate. This research suggests a remedy for the problem of scant data in training Tesseract LSTM for a new language by exploiting a training dataset for a language with a similar script. The target of the experiment is Kurdish and as a result, they have achieved a 95.45% accuracy rate.

## 3. Methodology

The section is divided into two subsections: the dataset description and the Method and Design.

## 3.1 Dataset Description

In the field of handwritten recognition, remarkable progress has been made over the past decade. Handwriting recognition research works in English, Chinese, and Arabic have been emphasized, and higher classification results were issued; nonetheless, very little research has been conducted on the Kurdish language. Although almost 30 million (Hassani, 2017) people worldwide speak Kurdish, for character recognition both online and offline, very few works have been accomplished and published in the direction of the automatic recognition of Kurdish scripts. This is a consequence of insufficient funding and other utilities, such as satisfactory datasets, dictionaries, etc.

**Table 1** number and percentage of the collected letters.

| Order | Letter | Number of images | Percentage |
|---|---|---|---|
| 1 | ئـ | 1134 | 2.77% |
| 2 | ا | 1134 | 2.77% |
| 3 | ب | 1134 | 2.77% |
| 4 | پ | 1008 | 2.46% |
| 5 | ت | 1134 | 2.77% |
| 6 | ج | 1134 | 2.77% |
| 7 | چ | 1260 | 3.08% |
| 8 | ح | 1260 | 3.08% |
| 9 | خ | 1134 | 2.77% |
| 10 | د | 1134 | 2.77% |
| 11 | ر | 1134 | 2.77% |
| 12 | ڕ | 1134 | 2.77% |
| 13 | ز | 1512 | 3.69% |
| 14 | ژ | 1123 | 2.74% |
| 15 | س | 1107 | 2.70% |
| 16 | ش | 1134 | 2.77% |
| 17 | ع | 1260 | 3.08% |
| 18 | غ | 1134 | 2.77% |
| 19 | ف | 1134 | 2.77% |
| 20 | ڤ | 1134 | 2.77% |
| 21 | ق | 1260 | 3.08% |
| 22 | ك/ک | 1386 | 3.39% |
| 23 | گ | 1134 | 2.77% |
| 24 | ل | 1134 | 2.77% |
| 25 | ڵ | 1134 | 2.77% |
| 26 | م | 1386 | 3.39% |
| 27 | ن | 1161 | 2.84% |
| 28 | هـ | 1008 | 2.46% |
| 29 | ە | 1512 | 3.69% |
| 30 | و | 1134 | 2.77% |
| 31 | ۆ | 1134 | 2.77% |
| 32 | وو | 1134 | 2.77% |
| 33 | ی | 1134 | 2.77% |
| 34 | ێ | 1134 | 2.77% |
| Total | | 40940 | 100% |

However, there seem to be no standardized datasets to be used as benchmarks in Kurdish (like the MNIST data set for English numbers). It is quite difficult to provide a comparable output of the existing methodology, and to the authors' knowledge, there seems to be no dataset of Kurdish Handwritten Characters available or published anywhere until this research is written.

Though the Kurdish language uses a very similar set of Arabic/Persian characters to write, and there are plenty of comprehensive Arabic, and Persian handwriting datasets for offline character recognition, None of these have all the characters used in Kurdish writing. For instance, (Mozaffari S, Faez K, Faradji F, Ziaratban M, Golzan S. M, 2006) built an extensive Isolated Farsi/Arabic character dataset containing 52,380 images of characters and 17,740 numerals. They have clarified that the dataset can also be used to recognize other languages, for example, Urdu and Kurdish, which use the same characters in writing. There are, however, two significant problems with this (Mozaffari S, Faez K, Faradji F, Ziaratban M, Golzan S. M, 2006) and other similar Datasets. Firstly, it doesn't include all the characters used in Kurdish, like Re (ر), Ve (ڤ), Le(ڵ), and Wo (ۆ).

The second problem is that it has no consistency in the number and percentage of characters. For example, there are some characters with 0.1 percent and others with 20 percent in the characters' total images. These two are the primary problems and will affect the system's accuracy and efficiency.

That's we built a comprehensive dataset of isolated Kurdish handwritten characters with 40,940 images of all 34 Kurdish characters. The dataset includes characters that have been replaced in recent years, like (ك), which is no longer being used by the Kurdish language (Hashemi, 2016) and has been replaced with (ک). Previously the dataset was published in a separate "data in brief" article (2021). The letters, number of images, and percentage of each character in the dataset are shown in Table 1. As the numbers indicate, it solves both problems mentioned earlier, thus, including all the characters and achieving a consistent percentage of all the characters.

## 3.2   Method and Design

Convolutional Neural Networks (CNN) is a class of deep models in machine learning, inspired by the information processing model in the human brain. In the human brain's visual part, each neuron has a receptive field capturing data in the visual space from a specific local neighborhood. They are designed specifically to recognize multidimensional data with a high degree of invariance to shift scaling and distortion. This section discusses a short overview of the CNN architecture, with an adequate explanation for every building block (Szegedy, Christian, et al, 2015).

The CNN network is composed of three stages of convolutions, three fully connected layers. Each convolution stage features a convolutional layer, a non-linear activation layer, and a max-pooling layer. The non-linear activation layer was not illustrated as the data size in those two layers was not changed. Figure 1 shows the overall CNN Architecture.

### 3.2.1 Layers

In this section, we will explain every layer of the architecture in the same order as Figure 1 showed. First, in subsection A, convolutional layers are presented. Then, in subsection B, the Non-linear activation layer is explained. In subsection C Pooling layer is covered, then in subsection D Flattening layer is analyzed. Next, in subsection E Output layer is defined, and finally, in subsection F Dropout strategy is explained.

- **Convolutional Layers**

In the context of the Convolutional Neural Networks architecture, a convolution operation implies that the local receptive field (kernel or filter) is reproduced to form a feature map throughout the entire field of vision. It shares the same (bias and weight vector) variables. Such weight-sharing reduces the number of free variables while increasing the network's performance. Weights (kernels or filters) are initialized as random and can be the color, edge, or particular pattern detectors.

- **Non-linear activation layer**

A convolution layer describes a window that functions as a linear filter. Therefore, non-linear activation functions that convert the input value to the expected value of the neuron are required to create a non-linear complex model. Initially, sigmoid activation functions were commonly used as an activation mechanism for neural networks.

The Sigmoid function can only operate between the minimum and maximum values; that's what causes the saturation problem. In addition, the activation function gradient in a saturated neuron is approximately zero, which could decrease the stream of gradients to the lower layers of the neural network.

ReLU was introduced by Glorot et al. (2011) as an alternative function and compared it with a softplus activation function (I Goodfellow, Y Bengio, A Courville, 2016).

- **Pooling Layers**

This section will briefly outline the pooling layer in CNN architecture. An output map with the local function is derived from prior linear convolution, adding up non-linear activation and bias. The observed characteristics can still be sensitive to the exact positions of the input pattern due to weight replication in a convolutional layer, which is damaging to output if subsequent classification is followed. An acceptable approach to solve the mentioned problem is to pool the features that have three main advantages; pooling efficiently to minimize dimensionality, which decreases feature map resolution, and computing of the Convolutional Neural Networks upper layer. It dumps out irrelevant data and preserves only the most important knowledge (Xiang Zhang, Junbo Zhao, Yann LeCun, 2015). Furthermore, pooling in a neural network makes activations less sensitive to the neural network's particular structure. (Nitish Srivastava et al. 2014).

The Pooling layer is employed to gradually reduce the number of parameters and computation in the CNN network, which operates individually on each feature map dimension (height and width).

- **Flatten Layers**

A (block) conversion stage is formed by hierarchically combining the four layers: convolutional layers, non-linear layer, pooling layer, and normalization layer. Then, it flattens the input image by passing it through various convolution stages to extract complex descriptive features in conventional CNN architecture. Finally, the flattened layer is employed to convert the feature maps of the multidimensional matrix into one-dimensional vectors. In addition, the flattened layer does not affect the batch size.

- **Output layer**

As the fully connected layers receive feature vectors from the topmost convolution stage, the output layer can generate a probability distribution over the output classes. Toward this purpose, the output of the last fully connected layer is fed to a K-way softmax layer (where K is the number of classes), which is the same as a multiclass logistic regression. If we denote by $x^i$, the $i^{th}$ input to the output layer, then the probability of the $i^{th}$ class, $p^i$, is calculated by the following softmax function.

$$p^i = \exp(x^i) / \sum_{j=1}^{K} exp\, x^j \qquad (1)$$

- **Dropout**

Efficient machine learning systems are deep neural networks with multiple parameters. Overfitting such networks, however, is a major concern. Extensive networks are often slow to use, and it isn't easy to handle overfitting by combining the predictions during testing of many different large neural networks. One of the ways to solve this dilemma, which is a problem-solving technique, is a dropout. The key idea is to randomly drop selected units from the neural network during planning (along with their connections), which prevents the units from being co-adapted too much. During training, dropout samples from an exponential number of various "thinned" networks. At test time, it is easy to approximate the effect of averaging the predictions of all these thinned networks by simply using a single unthinned network with smaller weights. This significantly reduces overfitting and gives substantial improvements and advantages over other regularization methods.

In dropout, we minimize the loss function stochastically under a noise distribution. This can be seen to reduce the predicted loss function. Previous work of Globerson and Roweis's (2006); Dekel et al. (2010) studied an alternate setting where the loss is minimized when an adversary gets to pick which units to drop. Here, rather than noise distribution, the maximum number of units that can be dropped is fixed. However, this work doesn't explore models with hidden units, though.

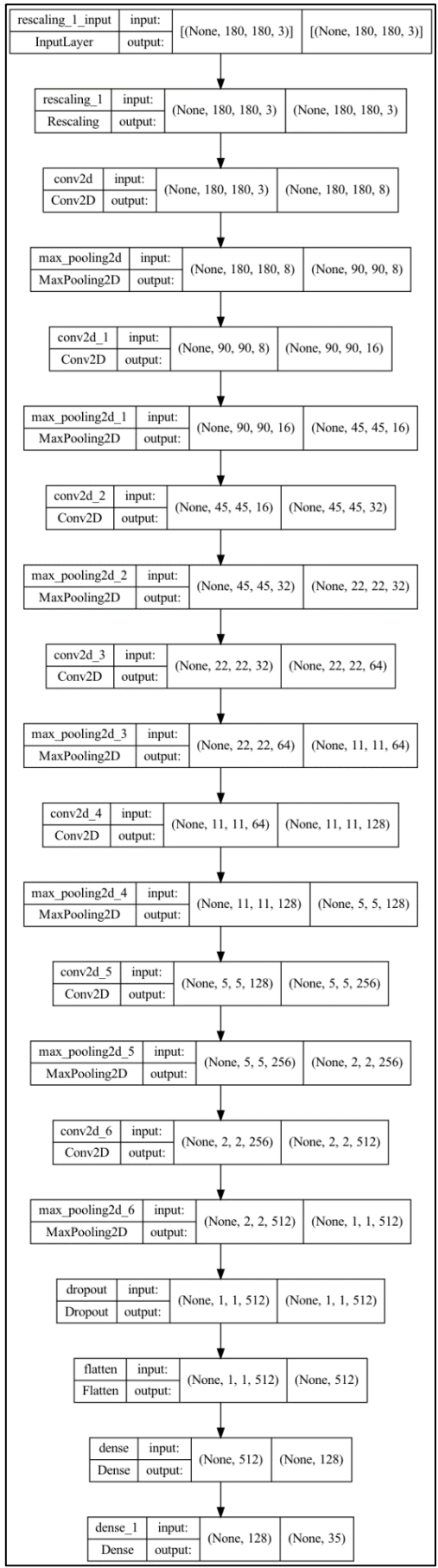

Figure 1: The CNN deep learning architecture for the Kurdish handwritten character recognition

# 4 Results

## 4.1 Training the Model

Here, we briefly outline the training procedure and strategies, then we will experiment with different parameters and plot the results.

The first step in training is loading the data. Our dataset is divided into three folders, one for training, validation, and testing. Each folder contains 35 subfolders for each letter, with a total of 40 thousand images.

Then the next part is Data Augmentation, which gets rid of overfitting. Usually, overfitting occurs once we have a few training data, so a way to address the problem is to increase our dataset to have an adequate number of training data. Data augmentation uses the idea of generating sufficient data from existing training samples, thus, increasing samples by multiple random modifications that produce credible images. The intention is your model will never see the same picture twice. This exposes the model to too many data aspects and makes it more general.

After that, we start creating our CNN Model using the Keras library in TensorFlow, which contains all the needed layers. First, we create a convolutional neural network of 3 convolution blocks. Each convolution block has a Conv2D layer and a max pool layer. The first convolutional block has 16 filters, the second one has 32 filters, and the third one has 64 filters. All convolutional filters are 3 x 3, and all max pool layers have a pool size of (2, 2).

After the three convolutional blocks, we have the flattened layer preceded by a fully connected layer with 512 units. CNN should output class probabilities based on five classes done by the **softmax** activation function. All other layers use the Relu activation function. Also, a dropout layer was added with a probability of 20%, where appropriate.

The CNN model was then compiled using the ADAM optimizer and sparse cross-entropy as a loss function. To see training and validation accuracy on each epoch as we train our network, we have printed them out for each epoch. After compiling the CNN Model, the model has been plotted to ensure that we have all the layers we need and that the data flow is as we designed. After the compilation step, a model summary has been printed out using Keras's summary() function, which prints all the details of all layers of the model, as shown in Figure 2.

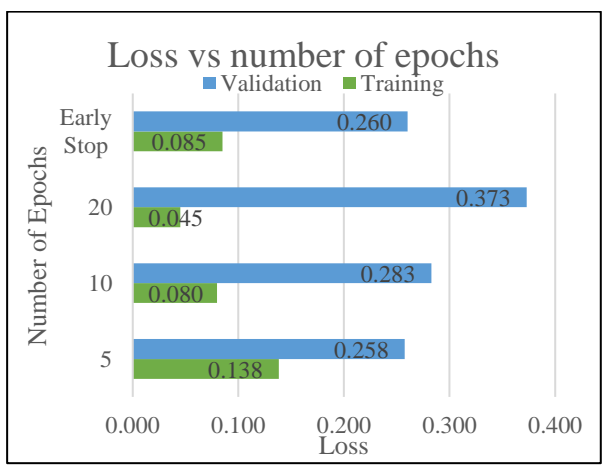

Figure 2 CNN Model Summary.

After setting all the hyperparameters, the training process starts. The network is trained with 5 epochs for the first experiment, and all the results are recorded. Then, after resetting all the parameters, the model was trained for 10 epochs, and all the results were recorded again. Finally, we trained the model with 50 epochs after resetting all the parameters.

From the output of the training process, we can get training and validation accuracy, and training and validation loss. After different training rates of the model, we can see that the higher the Epochs are, the better our accuracy, as shown in Figure 3.

Figure 3 Loss vs. Number of Epochs in 3 experiments.

## 4.2 Testing the Model

After all three experiments in section 4.1, we tested the model's accuracy with our testing dataset, which contains 8147 images of all the different characters, with an equal percentage of each character. For the testing, we did two separate tests for each experiment. First, the test was to give all the 8147 characters to the model to predict while we shuffled the data to be as random as possible, and the second test was to test each character individually. Figure 4 shows, training, validation, and testing accuracy versus number of epochs.

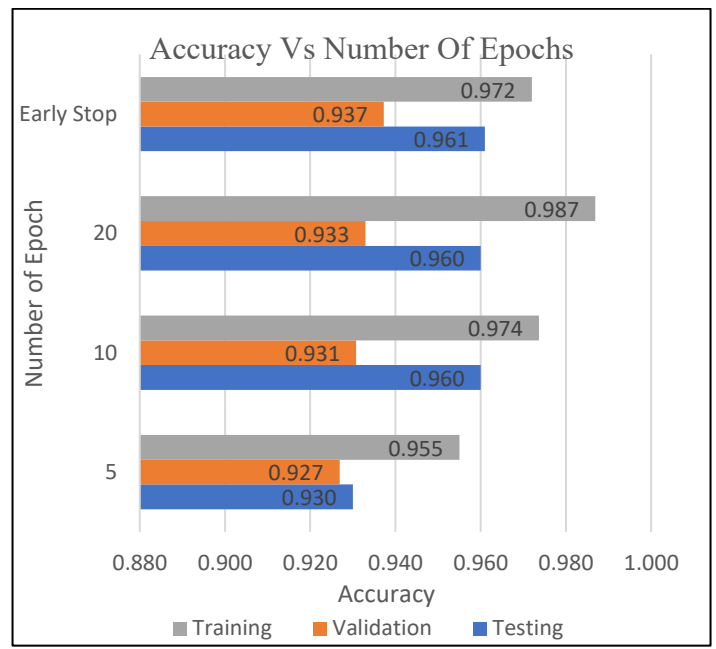

Figure 4 Training, Validation, and Testing Accuracy vs. Number of Epochs.

After the experiments were finished in section 4.1, we tried to analyze the errors that occurred during the testing time and decreased the model's accuracy. Throughout analyzing and sorting the false predictions, we found out that most false predictions are between similar characters in shape, such as (ﻚ) and (ﻝ), which are only different in one dot above. We found out that out of 1016 wrong predictions, 88 were the letter (ح) and the model predicted (خ); of course, they are very similar. 61 of them were the letter (ﻚ) and the model predicted (ﻝ), which again are very similar. Figure 5 shows the 10 most frequent misclassification in the testing.

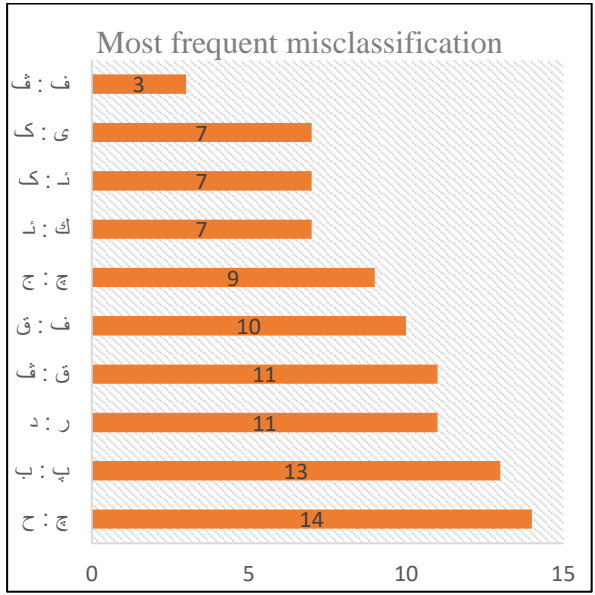

Figure 5 Most frequent misclassification in testing.

After finding the most frequent misclassification, as mentioned earlier in this section, then we shifted to testing each letter's accuracy separately. For each letter, we shuffled the folder of that letter and then fed it to the testing function. Figure 6 shows the result of the testing of individual letters.

## 4.3 Evaluation

throughout completing each iteration of the test, the results of our model are compared with the famous CNN Architecture AlexNet (Krizhevsky A, Sutskever I, Hinton, 2012), as shown in Figure 7, almost in all epochs, our model's accuracy surpasses the well-known AlexNet architecture.

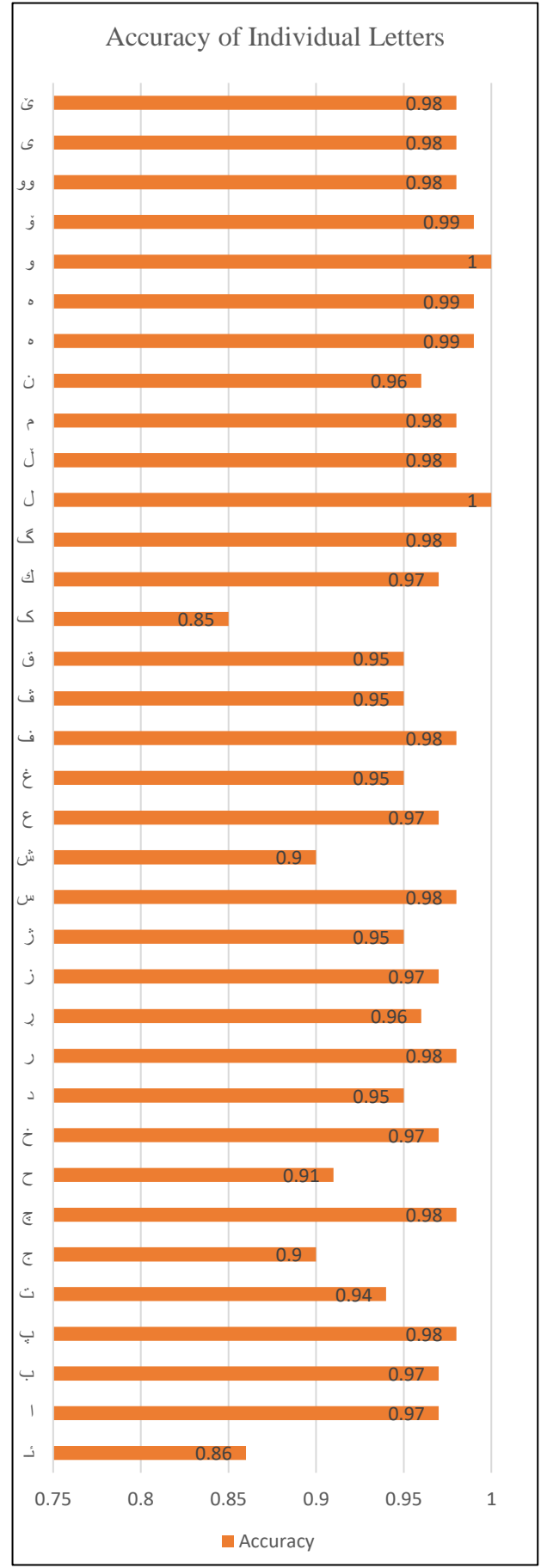

Figure 6, Accuracy of individual Letters.

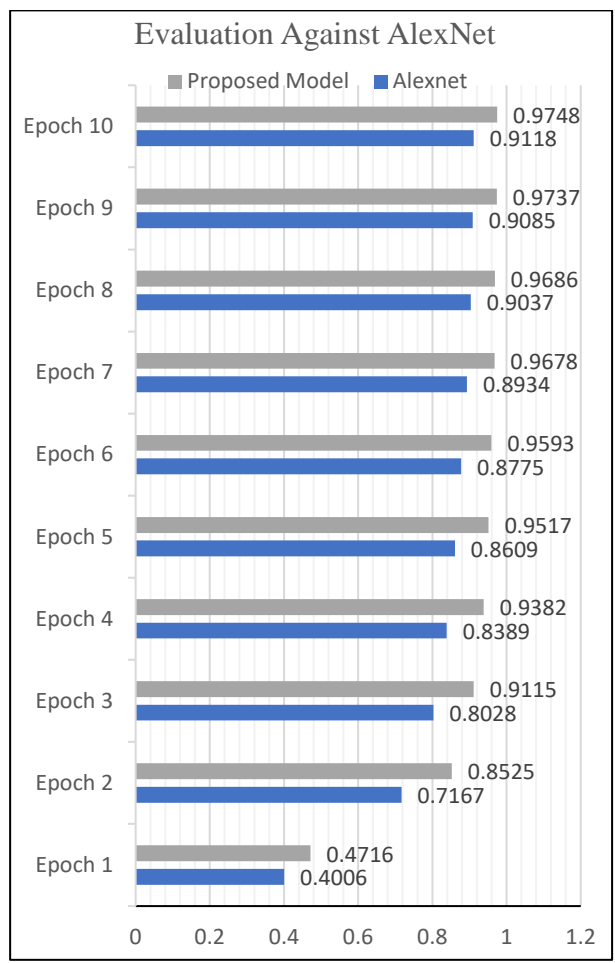

Figure 7 Evaluation with AlexNet.

# 5. Discussions

The study's primary purpose was to create a Kurdish handwritten character recognition. For this, we created a dataset and designed a CNN model to train the dataset. First, we trained the model over 60% of our dataset and validated it with 20%, which becomes 80% of the whole dataset. After that, we tested our CNN model with the remaining 20% of the dataset. Our results revealed %97 accuracies during training, 93% accuracy during validation, and %96 accuracies during the testing process; after training the model by 20 epochs, as shown in Figure 7, the more epoch we had, the more accurate the model became.

Therefore, some points need to be discussed about the training and testing process as listed below

- Overfitting: overfitting issue emerged on the version of the network that has been trained by 5 epochs, as shown in Figure 4. There is a difference between the training and validation accuracy results. The training accuracy is lower by 4% than validation accuracy, which is a sign of overfitting, one of the known issues of machine learning. On the contrary, the model goes underfitting when it is too simple regarding the data it is trying to model. In our case, the model is not too simple, and adding more data will not generally help. Our solution was to train the model with more epochs, and as it's summarized in Figure 4, it seems we don't have this problem in the network with 10 and 20 epochs.
- Evaluation Process: Evaluation results show that our model got higher accuracy on the dataset than AlexNet. Interestingly that doesn't mean our CNN model is better than AlexNet, even if it's performing better on this dataset since AlexNet was originally designed to classify the ImageNet, which is a much larger dataset that contains 14 million images.
- Misclassification: Most misclassification occurred within similar characters, as shown in Figure 5. The most misclassified letter is (چ), which has been classified as (ح). The letter (چ) essentially is the same character as (ح), but with added dots Inside. This issue resulted in very low accuracy in the testing as a result of all misclassifications. The next most misclassified letter was (ڤ), classified as (ڥ). Again, they have the same structure and with only one dot difference.
- Dataset: a very extensive dataset was created for isolated characters that contain more than one thousand images for each letter with very good quality. Still, all our writers were nearly at the same level of literacy and age. Our writers were either university students or academic staff, and the dataset only contained isolated characters, not all character shapes.

# 6. Conclusion

**Summary**: Handwritten character recognition is an application of artificial intelligence, computer vision, and pattern recognition. A computer with handwriting recognition can obtain and detect characters in paper documents, images, and touch screens. Researchers have used various techniques and tools over the years to solve handwriting recognition for different languages. Here, we studied deep learning on the problem because it has gained much consideration lately and is especially well suited for the type of learning that can potentially be very useful for real-world applications. For the implementation of this deep learning model, we chose TensorFlow, one of the trending frameworks for deep learning, and it has proven to be more powerful and easier to deploy than its competitors. Furthermore, we designed the model based on Convolutional Neural Networks. Finally, we trained the model with different parameters and rates and had different results for each type of setting and experiment.

**Conclusion:** Handwritten recognition is one of the major learning problems. This research presented a deep learning model to recognize Kurdish handwritten characters. Overall, our model showed a reasonable accuracy rate; in testing achieved a 96% accuracy rate and a 97% accuracy rate in training. On this basis, we concluded the research into the following points:

1) This research is aimed to collect a large amount of handwritten isolated Kurdish character images to create a dataset for evaluating our model and future research work in the field of Kurdish handwriting recognition.
2) The quality of the dataset in terms of examples of each letter has a substantial impact on the accuracy of the model because of the nature of Kurdish writing, which has a complicated structure. We have noted that the most frequent errors are between the strokes of similar characters like چ،ج and ڤ،ف.
3) Increasing the epochs of training will immediately affect the accuracy of the model. The more epochs we have, the more accurate the model will be because we have a large dataset and 35 class output, that's why the model needs to be trained more to cover more training samples.
4) Overfitting is one of the known problems for deep learning models. Adding dropout layers and early stopping, minimized the overfitting issue. As shown in Figure 4, validation and training accuracy are very close.

**Future Works:** Kurdish Handwritten recognition is in its first stages, this research presented an isolated handwritten character recognition, which can be considered as a building block for word recognition and full-page recognition. Next, the model's accuracy can be improved, or the dataset can be extended to contain all the shapes and forms of characters.

The desirable future work for this research could be the following:

1) Deep learning is a very hot topic right now. Many new techniques and tools can be applied as different feature extraction approaches and classifiers to recognize the characters and improve the model's accuracy.
2) Creating an extended dataset containing all the shapes and forms of Kurdish letters.
3) Creating a segmentation-based word recognition that first segments the letters and then passes them to the proposed model for recognition.

**Limitations and Challenges:** between the lack of datasets and standards in the Kurdish language, there are some limitations in any work related to the Kurdish language, and as for the implementation, we can conclude the limitations in the following points:

1. The achieved accuracy of 96% in testing is very reasonable and comparable to similar work in other languages, but it's not perfect, and especially if the model was used for a sensitive application, it needs to be better in terms of accuracy.
2. The proposed method and dataset are only focused on isolated characters, and to make it more usable and applicable, it can incorporate all the shapes of the characters, which wasn't feasible in the amount of time we had and was outside of the scope of the project.
3. The amount of overfitting that was appearing in the validation and testing accuracy was reduced with the help of early stopping and using dropout, but it didn't help the overfitting in the loss numbers as much.

**Ethics Statement**

All the handwriting was obtained with the consent of the individuals who had participated in the writing.

**Competing Interests**

The authors declare that they have no known competing financial interests or personal relationships which have, or could be perceived to have, influenced the work reported in this article.